\documentclass[journal]{IEEEtran}
\makeatletter\if@twocolumn\PassOptionsToPackage{switch}{lineno}\else\fi\makeatother

\usepackage{graphicx}
\usepackage[T1]{fontenc}
\ifCLASSINFOpdf
\else
\fi
\usepackage{amsmath}
\usepackage[cmintegrals]{newtxmath}
\usepackage{url,multirow,morefloats,floatflt,cancel,tfrupee}
\makeatletter

\AtBeginDocument{\@ifpackageloaded{textcomp}{}{\usepackage{textcomp}}}
\makeatother
\usepackage{colortbl}
\usepackage{xcolor}
\usepackage{pifont}
\usepackage[nointegrals]{wasysym}
\makeatletter

\def\mcWidth#1{\csname TY@F#1\endcsname+\tabcolsep}

\def\cAlignHack{\rightskip\@flushglue\leftskip\@flushglue\parindent\z@\parfillskip\z@skip}
\def\rAlignHack{\rightskip\z@skip\leftskip\@flushglue \parindent\z@\parfillskip\z@skip}

\@ifundefined{etal}{}{}

\usepackage{ifxetex}
\ifxetex\else\if@twocolumn\@ifpackageloaded{stfloats}{}{\usepackage{dblfloatfix}}\fi\fi

\AtBeginDocument{
\expandafter\ifx\csname eqalign\endcsname\relax
\def\eqalign#1{\null\vcenter{\def\\{\cr}\openup\jot\m@th
  \ialign{\strut$\displaystyle{##}$\hfil&$\displaystyle{{}##}$\hfil
      \crcr#1\crcr}}\,}
\fi
}

\AtBeginDocument{%
  \@ifpackageloaded{endfloat}%
   {\renewcommand\efloat@iwrite[1]{\immediate\expandafter\protected@write\csname efloat@post#1\endcsname{}}}{\newif\ifefloat@tables}%
}%

\def\BreakURLText#1{\@tfor\brk@tempa:=#1\do{\brk@tempa\hskip0pt}}
\let\lt=<
\let\gt=>
\def\processVert{\ifmmode|\else\textbar\fi}

\@ifundefined{subparagraph}{
\def\subparagraph{\@startsection{paragraph}{5}{2\parindent}{0ex plus 0.1ex minus 0.1ex}%
{0ex}{\normalfont\small\itshape}}%
}{}

\newcommand\role[1]{\unskip}
\newcommand\aucollab[1]{\unskip}
  
\@ifundefined{tsGraphicsScaleX}{\gdef\tsGraphicsScaleX{1}}{}
\@ifundefined{tsGraphicsScaleY}{\gdef\tsGraphicsScaleY{.9}}{}
\def\checkGraphicsWidth{\ifdim\Gin@nat@width>\linewidth
	\tsGraphicsScaleX\linewidth\else\Gin@nat@width\fi}

\def\checkGraphicsHeight{\ifdim\Gin@nat@height>.9\textheight
	\tsGraphicsScaleY\textheight\else\Gin@nat@height\fi}

\def\fixFloatSize#1{}
\let\ts@includegraphics\includegraphics

\def\inlinegraphic[#1]#2{{\edef\@tempa{#1}\edef\baseline@shift{\ifx\@tempa\@empty0\else#1\fi}\edef\tempZ{\the\numexpr(\numexpr(\baseline@shift*\f@size/100))}\protect\raisebox{\tempZ pt}{\ts@includegraphics{#2}}}}

\AtBeginDocument{\def\includegraphics{\@ifnextchar[{\ts@includegraphics}{\ts@includegraphics[width=\checkGraphicsWidth,height=\checkGraphicsHeight,keepaspectratio]}}}

\DeclareMathAlphabet{\mathpzc}{OT1}{pzc}{m}{it}

\def\URL#1#2{\@ifundefined{href}{#2}{\href{#1}{#2}}}

\def\UrlOrds{\do\*\do\-\do\~\do\'\do\"\do\-}%
\g@addto@macro{\UrlBreaks}{\UrlOrds}

\edef\fntEncoding{\f@encoding}

\makeatother

\newif\ifmultipleabstract\multipleabstractfalse%
\usepackage{tabulary}
\makeatletter
\AtBeginDocument{\@ifpackageloaded{longtable}{%
\def\LT@makecaption#1#2#3{%
  \LT@mcol\LT@cols c{\hbox to\z@{\hss\parbox[t]\LTcapwidth{%
    \sbox\@tempboxa{#1{#2: } #3}%
    \ifdim\wd\@tempboxa>\hsize
      #1{#2: }\textsc{#3}%
    \else
      \hbox to\hsize{\hfil\box\@tempboxa\hfil}%
    \fi
    \endgraf\vskip\baselineskip}%
  \hss}}}
}{}}
\makeatother

\makeatletter
\let\citep\cite
\let\citet\cite
\makeatother

   \makeatletter
  \def\fig@textbf{\textbf}
   \AtBeginDocument{\renewcommand\floatc@plain[2]{\setbox\@tempboxa\hbox{{\footnotesize#1.}\footnotesize\hskip.5em#2}%
    \ifdim\wd\@tempboxa>\hsize {\fig@textbf{\footnotesize#1.}}\footnotesize\hskip.5em#2\par
        \else\hbox to\hsize{\hfil\box\@tempboxa\hfil}\fi}}
    \makeatother
  
\usepackage[flushleft]{threeparttablex}

\usepackage{float}

\begin{document}

%

        \title{Emotion Based Hate Speech Detection using Multimodal Learning}
      
\author{Aneri~Rana and 
        Sonali~Jha}

\maketitle 

\begin{abstract}
 In recent years, monitoring hate speech and offensive language on social media platforms has become paramount due to its widespread usage among all age groups, races, and ethnicities. Consequently, there have been substantial research efforts towards automated detection of such content using Natural Language Processing (NLP). While successfully filtering textual data, no research has focused on detecting hateful content in multimedia data. With increased ease of data storage and the exponential growth of social media platforms, multimedia content proliferates the internet as much as the text data. Nevertheless, it escapes the automatic filtering systems. Hate speech and offensiveness can be detected in multimedia primarily via three modalities, i.e., visual, acoustic, and verbal. Our preliminary study concluded that the most essential features in classifying hate speech would be the speaker's emotional state and its influence on the spoken words, therefore limiting our current research to these modalities. This paper proposes the first multimodal deep learning framework to combine the auditory features representing emotion and the semantic features to detect hateful content. Our results demonstrate that incorporating emotional attributes leads to significant improvement over text-based models in detecting hateful multimedia content. This paper also presents a new Hate Speech Detection Video Dataset (HSDVD) collected for the purpose of multimodal learning as no such dataset exists today.
\end{abstract}
    

%
\IEEEpeerreviewmaketitle

\section{\textbf{Introduction}}
 Around the world, we are seeing a disturbing groundswell of racism and intolerance {\textemdash} including but not limited to rising antisemitism, anti-Muslim hatred, and hate crimes against Asians. Social media and other forms of communication are being exploited as platforms for bigotry. \textit{Hate Speech} is defined as any communication in speech, writing or behavior, that attacks or uses pejorative or discriminatory language with reference to a person or a group based on who they are; in other words, based on their religion, ethnicity, nationality, race, color, descent, gender or other identity factors. It can lead to incitement, which explicitly and deliberately aims at triggering discrimination, hostility, and violence\footnote{\protect\BreakURLText{https://www.un.org/en/genocideprevention/documents/UN\%20Strategy\%20and\%20Plan\%20of\%20Action\%20on\%20Hate\%20Speech\%2018\%20June\%20SYNOPSIS.pdf}}. Therefore, real-time detection and filtering of all forms of hate speech is paramount. 

Hate speech can be present in various formats like text, audio, images, and video on social media. Several NLP methods have been experimented with for hate speech detection in languages, such as neural networks\unskip~\cite{1298898:23920947,1298898:23920961,1298898:23920963}, n-grams\unskip~\cite{1298898:23920517,1298898:23920546}, and graph based models\unskip~\cite{1298898:23920971,1298898:23920972}. However, there is a lack of substantial research on multimedia data. Additionally, it should be noted that hate speech in multimedia data is not simply dependent on the text but the emotional effect that the tone and delivery of speech has on the end listener. According to Patrick et al.\unskip~\cite{1298898:23920576}, hate speech and offensive behavior are linked to the emotional and psychological state of the speaker, which is reflected in the effective emotions of their language\unskip~\cite{1298898:23920671}. For example, a political leader calmly speaking about immigration policies at a conference is less harmful than delivering the same speech with extreme anger and disgust towards the targeted immigrants, as the latter causes incitement against the immigrants in the country. Accounting for emotion in detecting hate speech also reduces the number of false positives detected by the systems that only consider text data as input, as emotion provides more context to the speaker's intent.

Therefore, we propose a method to classify hate speech using a multimodal deep learning architecture that combines semantic and emotion features extracted from the speech. We manually collected a \textit{hate speech detection video dataset }(HSDVD) as no such dataset exists today, to the best of our knowledge. Due to the limited size of the data, transfer learning\unskip~\cite{1298898:23920761} was employed to pre-train the models responsible for capturing the unimodal language and speech embedding. To summarize, there are three machine learning models. The first one detects hate speech in text and was built by pre-trainning transformer networks such as BERT and ALBERT on the existing Twitter data sets. The second one, or the emotion detection \textit{multi-task learning} (MTL) model, is trained on the IEMOCAP dataset to detect the level of valence, arousal, and dominance in the audio. Due to the challenge of discrete representation of complex emotions\unskip~\cite{1298898:23920764} such as anger or fear, we use a dimensional representation of emotions defined by valence, arousal, and dominance attributes\unskip~\cite{1298898:23920765}.

Lastly, for multimodal learning (MML), a multilayer perceptron model is trained on HSDVD to detect hate speech. It selects the best performing text and emotion models to generate embeddings  as input. In addition, the text based models are fine-tuned and evaluated on HSDVD to create a benchmark for comparison with the MML framework. Both are tested on a holdout dataset from HSDVD. We experimented with two baselines, and the MML framework exemplified a gain in precision and recall across its respective baselines. This confirms our hypothesis on the significance of multiple modalities for the hate speech detection task.
    
\section{\textbf{Literature Review}}
Research on multimedia hate speech content has begun to emerge in the last couple of years, yet the prior work is limited. The hateful memes challenge by Facebook drew significant attention from researchers in 2020\footnote{\protect\BreakURLText{https://ai.facebook.com/blog/hateful-memes-challenge-winners/}}. The top 3 winning teams used pre-trained multimodal transformer models to combine the visual features of the image with textual features of the caption with considerable success\unskip~\cite{1298898:23921024,1298898:23921026,1298898:23921027}.

A much recent effort was also made in the field of video hate speech detection\unskip~\cite{1298898:23921029}. However, it only focuses on the text aspect of the video discarding any additional features which multimedia data provides. Another research focusing on offensive video detection collected and published a dataset in Portuguese\unskip~\cite{1298898:23921030}. It used transcripts along with social media features like tags and titles to detect offensiveness. Nevertheless, such models are dependent on features available after the content has reached a wider audience, rendering damage to the targeted group in the process. Therefore, a need for a more sophisticated method of hate speech detection in multimedia data arises.

To perform feature extraction on text, we train a hate speech detection language model. The NLP techniques for hate speech detection have evolved through several stages in the past decade. Early approaches experimented with TFIDF\unskip~\cite{1298898:23920546}, bag-of-words (BOW) or n-gram\unskip~\cite{1298898:23920517}, user-specific features such as age or social media features like shares, retweets or reporting for author profiling\unskip~\cite{1298898:23920971,1298898:23921061}. Since the maturity of deep learning approaches, much of the research in recent years has focused on neural architectures. Badjatiya et al.\unskip~\cite{1298898:23920560} and Gamback et al.\unskip~\cite{1298898:23920573} were the first to use recurrent neural networks (RNNs) and convolution neural networks (CNNs), respectively, for hate speech detection in tweets. Current state-of-the-art models fine-tune pre-trained transformers like BERT and ALBERT. Seven of the top ten teams in offensive language detection tasks (2019) used BERT with some variation in parameters and pre-processing\unskip~\cite{1298898:23921110,1298898:23921127}. In 2020 again\unskip~\cite{1298898:23921111}, the top ten teams used combinations of BERT, RoBERTa, or XML-RoBERTa, and the winning team used ALBERT\unskip~\cite{1298898:23921680}.

In Speech Emotion Recognition (SER), there has been extensive research, as it has a wide variety of applications such as Human-Computer Interaction\unskip~\cite{1298898:23921161}, Sentiment Analysis, and Enhancing film sound design\unskip~\cite{1298898:23921192}. Researches can be divided into two categories based on whether they classify the speech into an emotional state (happiness, sadness, anger, fear, disgust, boredom)\unskip~\cite{1298898:23921202} or predict the emotional attributes, that are, valence, arousal, and dominance\unskip~\cite{1298898:23921203}. Zisad et al.\unskip~\cite{1298898:23921230} has built a CNN model on the Ryerson Audio-Visual Database of Emotional Speech and Song (RAVDESS) data merged with some locally generated problem specific data collected by them to classify the speech emotion as calm, angry, fearful, disgust, happy, surprise, neutral or sad. The research uses tonal properties like MFCCs as features for the model. Zhang et al.\unskip~\cite{1298898:23921234} proposed that an attention-based CNN model trained on the speech spectrogram as the input instead of acoustic or statistical features can give better results. It is to be noted that the research also focuses on archetype emotion classification. On similar notes, Weiser et al.\unskip~\cite{1298898:23921237} compares an end-to-end learning network trained on raw audio data with a feature based network. Bojani\'{c} et al.\unskip~\cite{1298898:23921241}, in their call center based SER system focus on both archetype emotions as well as the emotional attributes of speech. Parthasarathy and Busso\unskip~\cite{1298898:23921242}, in their research, claim that the emotional attributes are interrelated and hence predicting them with a unified learning framework will give us better results. They make use of Multi-task learning along with Deep Neural Network models. 
    
\section{\textbf{Datasets}}
In this transfer learning approach, the target domain's feature space must overlap with the source domain to ensure a positive transfer\unskip~\cite{1298898:23920761}. Additionally, the consistency of training data distribution in all the models is critical to the success of transfer learning\unskip~\cite{1298898:23921613}. Hence, we experiment with multiple hate speech detection datasets for the text models and an acoustic dataset consisting of a varying range of emotional attributes for the emotion model.

The growing interest in automated hate speech detection has led to the creation of a plethora of text datasets from sources like Yahoo, Twitter, Wikipedia comments, and Reddit\unskip~\cite{1298898:23921629}. Twitter datasets were selected for this task, as the microblogging platform is closest to the target multimedia domain consisting of video/audio blogs. Each dataset has its advantages but also suffers from unintended biases\unskip~\cite{1298898:23921630}. Therefore, multiple combinations of datasets were experimented with, and the best model was chosen for feature extraction. All the datasets were subjected to the same pre-processing steps, i.e., removing Twitter user handles, URLs, and special characters (except exclamation and question marks).

\textbf{OffensEval 2019:} This dataset by Zampieri et al.\unskip~\cite{1298898:23921633} has been used extensively in SemEval 2019 Task 6\unskip~\cite{1298898:23921110} and SemEval 2020 Task 12\unskip~\cite{1298898:23921111} with considerable success. The data was collected from Twitter with the help of keywords and was annotated using crowdsourcing. It consists of 13k tweets and three levels of annotation. We focus only on level A annotations classifying it as offensive (33\%) or not and level B annotations classifying the targeted offense towards an individual (17.8\%), a group (8.2\%), or otherwise. Posts containing profane language and targeted offense, including insults, threats, or swear words, were classified as offensive.

\textbf{Waseem and Hovy 2016 (W\&H):} A dataset of 16k TweetIDs was published by Waseem and Hovy\unskip~\cite{1298898:23921661}. The authors annotated the tweets into three classes, namely racism (11.7\%), sexism (20.0\%), or neither. An additional third-party review recorded an inter-annotator agreement of 0.84. Only 10k tweets were retrieved, as the remaining had been taken down since 2016. The class distribution of the retrieved dataset is 9.2\% racism, 17.5\% sexism, and 73.3\% neither. It was noted by Madukwe et al.\unskip~\cite{1298898:23921629} that the dataset might be biased towards specific users since all the racist tweets were collected from only nine users. Mishra et al.\unskip~\cite{1298898:23921718} also called attention to specific tweets that lacked explicit abusive traits but were annotated as racist or sexist regardless.

\textbf{Davidson 2017:} Davidson et al.\unskip~\cite{1298898:23921672} published a dataset of 24k tweets, which were labeled as hate speech (5.77\%), offensive (77.43\%), or neither. Tweets were collected using a lexicon compiled by hatebase.org\footnote{\protect\BreakURLText{https://hatebase.org/}} containing hateful words and phrases. It was further annotated by CrowdFlower workers (now known as Appen\footnote{\protect\BreakURLText{https://appen.com/}}), and an inter-annotator agreement of 0.92 was reported. Madukwe et al.\unskip~\cite{1298898:23921629} noted inconsistencies in the labels and the lack of a diverse group of annotations in this dataset.

A combination of the aforementioned datasets was used to create three subsets of data for the text model experiments, as shown in Table~\ref{tw-99bbdca0d183}. OffensEval being the most reliable data source was used in dataset A. For dataset B, W\&H was combined with a subset of the Davidson dataset. To avoid class imbalance, Davidson was sub-sampled to retain only 3.5k offensive tweets. The racist tweets from the W\&H dataset were dropped in dataset C to further reduce noisy data.

\begin{table}[!htbp]
\caption{{Class Distribution of Twitter Datasets} }
\label{tw-99bbdca0d183}
\def\arraystretch{1}
\ignorespaces 
\centering 
\begin{tabulary}{\linewidth}{p{\dimexpr.0996\linewidth-2\tabcolsep}p{\dimexpr.33599999999999994\linewidth-2\tabcolsep}p{\dimexpr.27620000000000005\linewidth-2\tabcolsep}p{\dimexpr.2882\linewidth-2\tabcolsep}}
\hline \multicolumn{2}{p{\dimexpr(.43559999999999995\linewidth-2\tabcolsep)}}{Dataset} & Hatespeech & Not Hatespeech\\
\hline 
A &
  OffensEval &
  4400 (33.2\%) &
  8840 (66.8\%)\\
B &
   Davidson + W\&H &
  8408 (41.5\%)  &
  11839 (58.4\%)\\
C &
  Davidson + W\&H (no racist tweets) &
  5620 (32\%) &
  11963 (68\%)\\
\hline 
\end{tabulary}\par 
\end{table}
\textbf{IEMOCAP Dataset 2007: }The interactive emotional dyadic motion capture database, collected by the Speech Analysis and Interpretation Laboratory (SAIL) at the University of Southern California, is an audio-visual database\unskip~\cite{1298898:23921714}. A total of ten actors were recorded while performing a given script, but to maintain some originality in the emotions they were also asked to improvise certain hypothetical scenarios. It has approximately 12 hours of recordings. In total, the corpus contained 10039 samples (scripted session: 5255 samples; spontaneous sessions: 4784 samples) of an average duration of 4.5 seconds. Annotators were asked to evaluate the corpus in terms of the attributes valence (1-negative, 5-positive), arousal (1-calm, 5-excited), and dominance (1-weak, 5-strong). Two different annotators were asked to evaluate a single sample, and then speaker dependent z-normalization was used to compensate for inter-evaluator variation. 

\textbf{Hate Speech Detection Video Dataset (HSDVD): }Multimedia data consists of various types of media like text, images, audio, video, and animation. No existing hate speech dataset currently incorporates video and audio data, which accounts for a significant amount of content produced on social media platforms today. For this purpose, we compiled the Hate Speech Detection Video Dataset (HSDVD). It was collected from Twitter and YouTube using Twitter API and PyTube, respectively. Hate speech comprises abuses targeting a broad category of ethnicity, gender, or other identities. Since HSDVD will be used only to compare relative performances and due to the availability of limited resources, we collected data concerning specific groups so that each group has sufficient training data for the model. We began by creating a lexicon of sexist and ethnic slurs\footnote{\protect\BreakURLText{https://en.wikipedia.org/wiki/List\_of\_ethnic\_slurs\_by\_ethnicity}} focusing on these groups of people. Tweets with video posts were searched using the lexicon and Twitter API. On the other hand, YouTube videos primarily consist of viral social media posts when original copies were taken down. Therefore, we searched by combining the lexicon with phrases like \textit{went viral or racist rant}. Finally, 1k records of Tweets IDs and YouTube links were collected and classified as hate speech (25\%) or not. Both authors labeled each record upon pre-deciding on the definition of hate speech and the target groups. We computed the \textit{Cohen's kappa coefficient} of 0.92, indicating a high agreement.

The collected dataset focuses on hate speech targeted towards a gender, sexual orientation, autistic minority, Muslim, Jews, Sikh, Latino, Native Americans, and Asians. Although various definitions of hate speech exist, the common elements we considered are\unskip~\cite{1298898:23921783} :

\begin{itemize}
  \item \relax use of sexist, ethnic, or racial slurs intended to incite violence or hate against a target group
  \item \relax intended to threaten or harm a target group
  \item \relax abusive or offensive words being used to attack a target group
  \item \relax humor being used to attack a target group
  \item \relax intended to be derogatory or humiliating to a target group
  \item \relax negative stereotyping of a target group
\end{itemize}
  It is important to note that some terms are offensive when used to hurt a target group but can also be used casually by the same group, such as n**ga. Other terms can be used abusively but also in raps or among friends such as f**k and bi**h. Almost all terms that can be used to cause offense, can also be used for informational purposes to spread awareness. Moreover, some videos can be implicitly offensive without containing any explicit slurs. Such nuances are often difficult to differentiate using machine learning models. However, it could be more easily distinguishable in audio/video where emotions and intent are obvious, unlike text content, hence the model can benefit from both audio and text modalities.

According to Poletto et al.\unskip~\cite{1298898:23921630}, most of the existing hate speech datasets might be prone to unintended annotator bias and topic bias, including the above mentioned text datasets and HSDVD. However, our purpose is to demonstrate the effect of multiple modalities when detecting hate speech in multimedia and compare only the models' relative performances. Hence, our results will most likely to be independent of biases that may exist in the annotations.
    
\section{\textbf{Approach}}
We propose three deep learning models, namely text based hate speech classification, speech based emotion attribute prediction, and finally, a multimodal deep learning model to classify hate speech based on text and emotion.

Modality refers to a way in which something is experienced or expressed. Humans perceive the world around us through multiple modalities such as sound, image, language, smell, and taste. In the context of hate speech, an individual determines something as racist or sexist after combining visual, audio, and textual information from a particular instance. Therefore, a progressive artificial intelligence system should also consider the information in different formats while determining hate speech. In this paper, we focus on language, which captures the semantic information and vocal signal that encodes paraverbal information through the tone, pitch, and pacing of the voice. Although visual features have the potential to provide additional perspectives to hate speech detection, its efficacy was finite when learning over the limited scope of the current dataset. For example, only 15\% of the dataset contains distinguishable facial expressions, while in 10\% of the videos, religious objects could be linked to hate speech comments. Therefore, an extensive analysis with additional resources and extending HSDVD can be highly beneficial when included in future work. Nevertheless, with the current findings, we hope to build the foundation and observe the advantages of these two modalities compared to one in the multimedia hate speech detection task.

The most critical aspect for the performance of multimodal learning specific to this task is the representation and fusion of information for prediction, which can be achieved together using deep neural networks. We use a \textit{joint representation} technique that combines unimodal signals into a common representation space. There are other kinds of techniques called coordinated representations that process unimodal signals separately but under certain similarity constraints to project them onto a coordinated space\unskip~\cite{1298898:23983452}. The joint technique is used for this task, as all the modalities are present during both training and inference steps.  Mathematically, it is expressed as:
\let\saveeqnno\theequation
\let\savefrac\frac
\def\dispfrac{\displaystyle\savefrac}
\begin{eqnarray}
\let\frac\dispfrac
\gdef\theequation{1}
\let\theHequation\theequation
\label{dfg-10c1ba7e0370}
\begin{array}{@{}l}x_m\;=\;f(x_1,\;x_2,\;...,\;x_n)\end{array}
\end{eqnarray}
\global\let\theequation\saveeqnno
\addtocounter{equation}{-1}\ignorespaces 
where the multimodal representation $x_m $ is calculated using a function $f $ (a neural network activation function) that combines unimodal representations $x_1,\;x_2,\;...,\;x_n $. 

For unimodal representations of text and audio data, deep neural networks have become increasingly effective in the last decade\unskip~\cite{1298898:23922297,1298898:23922298,1298898:23922300}. We train two separate neural networks to learn text and audio embeddings. Since each successive layer in deep learning captures more abstract features, we use the last layer for final representations\unskip~\cite{1298898:23922306}. To construct the multimodal representation, the individual embeddings of both modalities are then fed to a common neural layer that projects these modalities into a joint space\unskip~\cite{1298898:23922308,1298898:23922309}. Multiple hidden layers can be used for further training from the multimodal representations before a non-linear classification layer is added to make the predictions (as seen in Figure~\ref{f-00ad823cc8aa}).

As neural networks require a large amount of labeled training data, it is common to pre-train embeddings. Since HSDVD contains only 1000 records, we utilize the technique of \textit{transfer learning }to train the unimodal embeddings. Transfer learning helps in improving the performance of a model on the target domain by transferring the knowledge from a different but related source domain\unskip~\cite{1298898:23920761}. In this approach of homogeneous transfer learning \textit{X}\ensuremath{_{S}} = \textit{X}\ensuremath{_{T}}, where \textit{X }is the space of all possible feature vectors in a domain \textit{D}\unskip~\cite{1298898:23922312}. As previously mentioned, this type of transfer learning is sensitive to differences in the domain and distribution of data. Therefore, the main objective is to reduce the distribution difference between source and target domains\unskip~\cite{1298898:23921613}. In the pre-trained text model, both the input feature space and labels extensively overlap with that of the target domain. Whereas for the speech model, the feature space is the same, and the classification objectives are different.

\subsection{\textbf{Text Based Hate Speech Classification Task }}In order to obtain text embeddings, we experiment with transformer networks, i.e., \textit{BERT base uncased} and \textit{ALBERT base uncased}, pre-trained on offensive tweets from different datasets A, B, and C to classify hate speech in text. These models are further fine-tuned and evaluated on HSDVD to serve as the baseline for comparing the performance of multimodal over text based learning. BERT stands for Bidirectional Encoder Representations from Transformers\unskip~\cite{1298898:23922403} and its architecture has been proven highly successful for transfer learning\unskip~\cite{1298898:23921680}. It was pre-trained on a 3.3 Billion text corpus and can be fine-tuned on any text based classification task. ALBERT or A Lite BERT, is also a successor of BERT that improves upon its architecture for memory efficient and faster training\unskip~\cite{1298898:23922404}. 

The tweets from all the datasets are subjected to the same pre-processing steps as specified in section 3. The sentences are broken down into token sequences of fixed length using WordPiece embeddings\unskip~\cite{1298898:23922510}. A special token [CLS] is also appended to the beginning of each sequence. After model training, we experiment with two methods of extracting text embeddings. The final text feature representation to be used for MML can be given by $E_t=e_{t1},\;e_{t2},\;\dots,\;e_{tl} $, where $l $ is the size of the embedding.

\subsection{\textbf{Emotion Attribute Prediction Task}}Unlike pre-trained transformers for text, we build out the multi-task deep learning model for emotion attribute prediction. The array of human emotions can be represented on a three-dimensional feature space defined by the attributes called valence, arousal, and dominance. Given the complexity of human interactions\unskip~\cite{1298898:23922651}, it is better to represent emotions using these abstract attributes instead of discrete classes such as happy, angry, or sad\unskip~\cite{1298898:23920764}. Moreover, due to the interrelation between these attributes\unskip~\cite{1298898:23921242}, a unified learning framework gives better results compared to single-task learning. The success of the multi-task learning paradigm has also been illustrated in the MTL survey by Zhang et al.\unskip~\cite{1298898:23922663}. Therefore, we formulate the prediction of each emotion attribute ranging between 0 and 1 as a regression problem solved using a multi-task deep learning model. The proposed approach jointly learns the objective of predicting valence, arousal, and dominance as continuous values (Figure~\ref{f-87efa4b75f31}).

The input representation for the IEMOCAP audio data are speech features including energy, entropy, chroma, spectral and mel frequency cepstral coefficient (MFCC) features computed using the pyAudioAnalysis\footnote{\protect\BreakURLText{https://github.com/tyiannak/pyAudioAnalysis}} feature extractor. It extracts frame by frame short term features with a fixed window size, where the frame moves over the speech signal one step size at a time\unskip~\cite{1298898:23922777}.  Over each short term feature sequence, statistical calculations such as arithmetic mean and standard deviation are performed to extract mid term features defined by their window size and step size parameters. We experiment with two different input representations using these feature vectors, which are then fed to the neural network.

\bgroup
\fixFloatSize{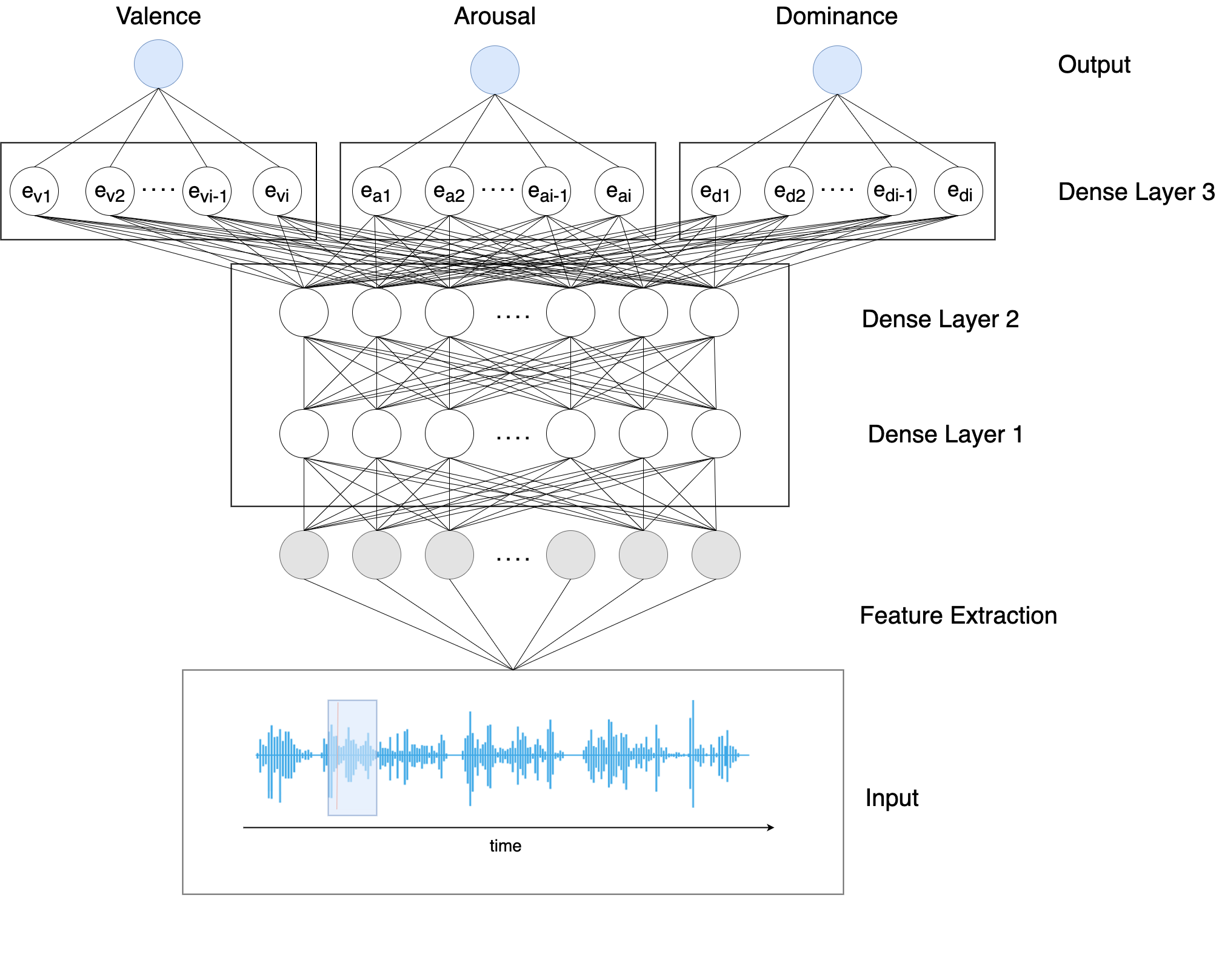}
\begin{figure}[!htbp]
\centering \makeatletter\IfFileExists{images/c50b9b8b-6c44-4e74-afc7-623d92bf00ea-uemotion_model-udrawio.png}{\includegraphics{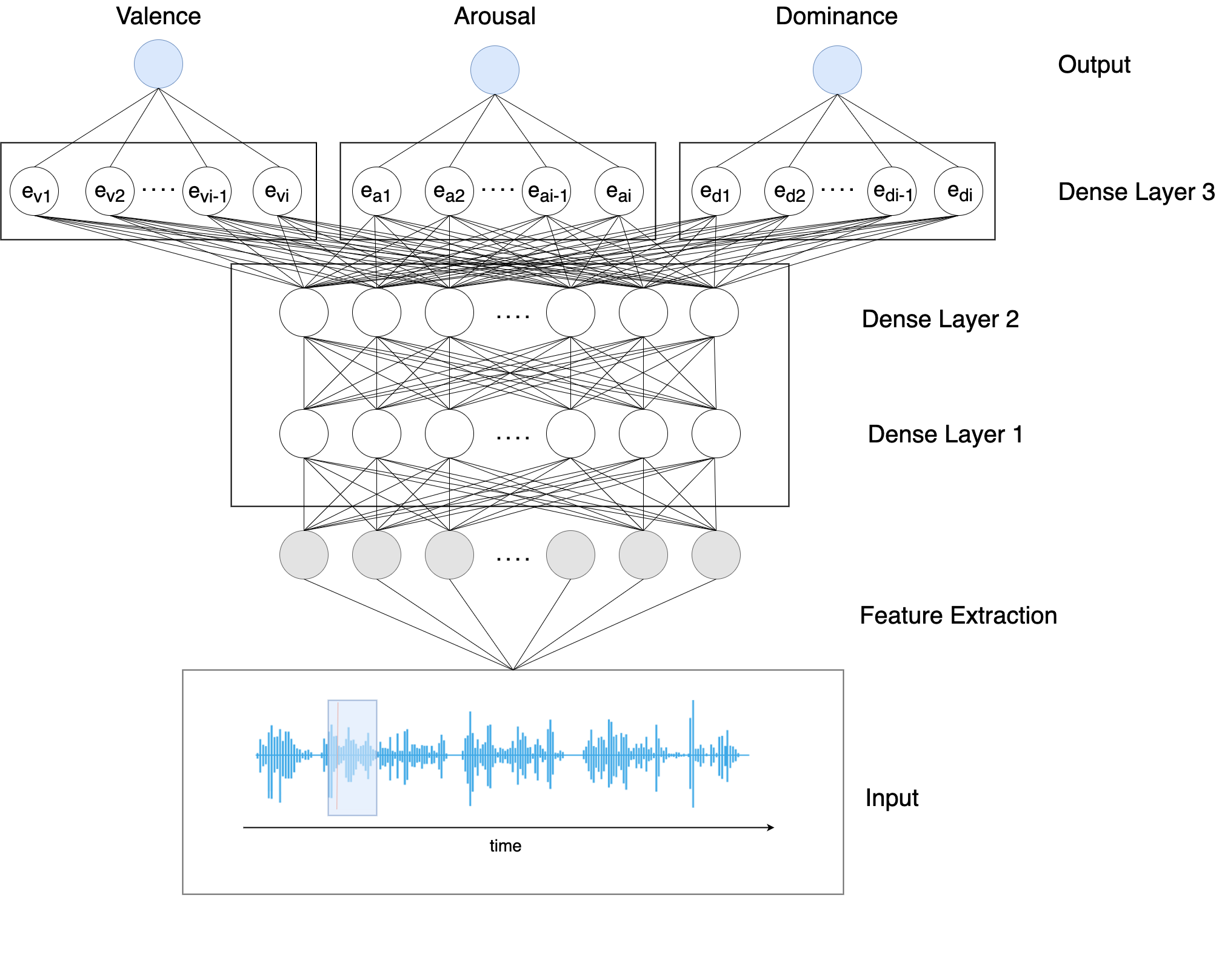}}{}
\makeatother 
\caption{{Multi-task learning for predicting valence, arousal and dominance. Dense Layer 3 shows the speech embeddings generated for MML.}}
\label{f-87efa4b75f31}
\end{figure}
\egroup
The MTL framework inspired by Caruana et al.\unskip~\cite{1298898:23922778} and depicted in Figure~\ref{f-87efa4b75f31} uses hard parameter sharing. It consists of two neural layers that are shared among all the attributes. These shared nodes create a joint representation for valence, arousal, and dominance. The third layer however, is specific to each task and learns representations to optimize that task. The entire model is trained to minimize the Mean Square Error (MSE) loss. This will generate three loss functions for each attribute, given by, $L_{val},\;L_{ars},\;L_{dom} $. The overall loss $L_{o} $ is calculated by taking a weighted sum of all the losses. Equation~(\ref{dfg-6d60c6ff049c}) gives the overall loss function for the MTL model.
\let\saveeqnno\theequation
\let\savefrac\frac
\def\dispfrac{\displaystyle\savefrac}
\begin{eqnarray}
\let\frac\dispfrac
\gdef\theequation{2}
\let\theHequation\theequation
\label{dfg-6d60c6ff049c}
\begin{array}{@{}l}L_o\;=\;\alpha\ast Lval\;+\;\beta\ast Lars\;+\;\gamma\ast\;Ldom\;\end{array}
\end{eqnarray}
\global\let\theequation\saveeqnno
\addtocounter{equation}{-1}\ignorespaces 
where the values of $\alpha,\;\beta $ and $\gamma $ vary between 0 and 1, in the steps of 0.1, such that $\alpha+\beta+\gamma\leq1 $. These hyper parameters are tuned using a validation set during the experiments. Finally, the attribute specific layers are used to create speech embeddings given by Equation~(\ref{dfg-74fb9942c041}).
\let\saveeqnno\theequation
\let\savefrac\frac
\def\dispfrac{\displaystyle\savefrac}
\begin{eqnarray}
\let\frac\dispfrac
\gdef\theequation{3}
\let\theHequation\theequation
\label{dfg-74fb9942c041}
\begin{array}{@{}l}E_s\;=\;concat(E_v,\;E_a,\;E_d)\;=\;e_{s1},\;e_{s2},\;\dots\;,\;e_{sm}\end{array}
\end{eqnarray}
\global\let\theequation\saveeqnno
\addtocounter{equation}{-1}\ignorespaces 
where  $E_v,\;E_a $ and $E_d $ are feature vectors generated by the respective Valence, Arousal, and Dominance specific layers.

\subsection{\textbf{Mutli-Modal Hate Speech Classification Task}}Figure~\ref{f-00ad823cc8aa} shows the multimodal learning (MML) framework where both text and audio embeddings are fed to three dense layers, followed by a classification layer to classify it as hate speech or not. As specified in section 4, a neural network layer can be used to combine the unimodal embeddings where each node activation function projects multiple modalities onto a shared space. The input can be represented as:
\let\saveeqnno\theequation
\let\savefrac\frac
\def\dispfrac{\displaystyle\savefrac}
\begin{eqnarray}
\let\frac\dispfrac
\gdef\theequation{4}
\let\theHequation\theequation
\label{dfg-6299b0b1dd01}
\begin{array}{@{}l}E\;=\;concat(E_t\;,\;E_s)\;=\;e_1,e_2,\;\dots,\;e_n\end{array}
\end{eqnarray}
\global\let\theequation\saveeqnno
\addtocounter{equation}{-1}\ignorespaces 
where $n=l+m $, such that $E_t\;\in\;\mathbb{R}^{l} $ and $E_s\;\in\;\mathbb{R}^{m} $. This is followed by three multi perceptron layers interleaved with two dropout layers to prevent overfitting\unskip~\cite{1298898:23922876}. Each of the hidden dense layer uses a rectified linear unit (ReLU) activation function\unskip~\cite{1298898:23922895} defined by $f(x)\;=\;max(0.0,\;x) $. It increases the convergence of stochastic gradient descent and has been proven to overcome the shortcomings of other functions such as Tanh and Sigmoid\unskip~\cite{1298898:23922896}. For weight initialization in these layers, He normalized initialization\unskip~\cite{1298898:23922897} is applied. It is a minor adaption of the Xavier normalized initialization to address specific characteristics of the ReLU activation function, which is non-linear for half of its input\unskip~\cite{1298898:23922898}. Finally, a sigmoid activation function is used at the output layer to predict $p(y) $ given by:
\let\saveeqnno\theequation
\let\savefrac\frac
\def\dispfrac{\displaystyle\savefrac}
\begin{eqnarray}
\let\frac\dispfrac
\gdef\theequation{5}
\let\theHequation\theequation
\label{dfg-d360fc803c2a}
\begin{array}{@{}l}p(y)\;=\;\sigma (x)\;=\;\frac1{1+e^{-x}}\;\;\end{array}
\end{eqnarray}
\global\let\theequation\saveeqnno
\addtocounter{equation}{-1}\ignorespaces 
where $p(y) $ generates a value between 0 and 1. A threshold of 0.7 is used to make the binary prediction on whether an input is hate speech or not.

\bgroup
\fixFloatSize{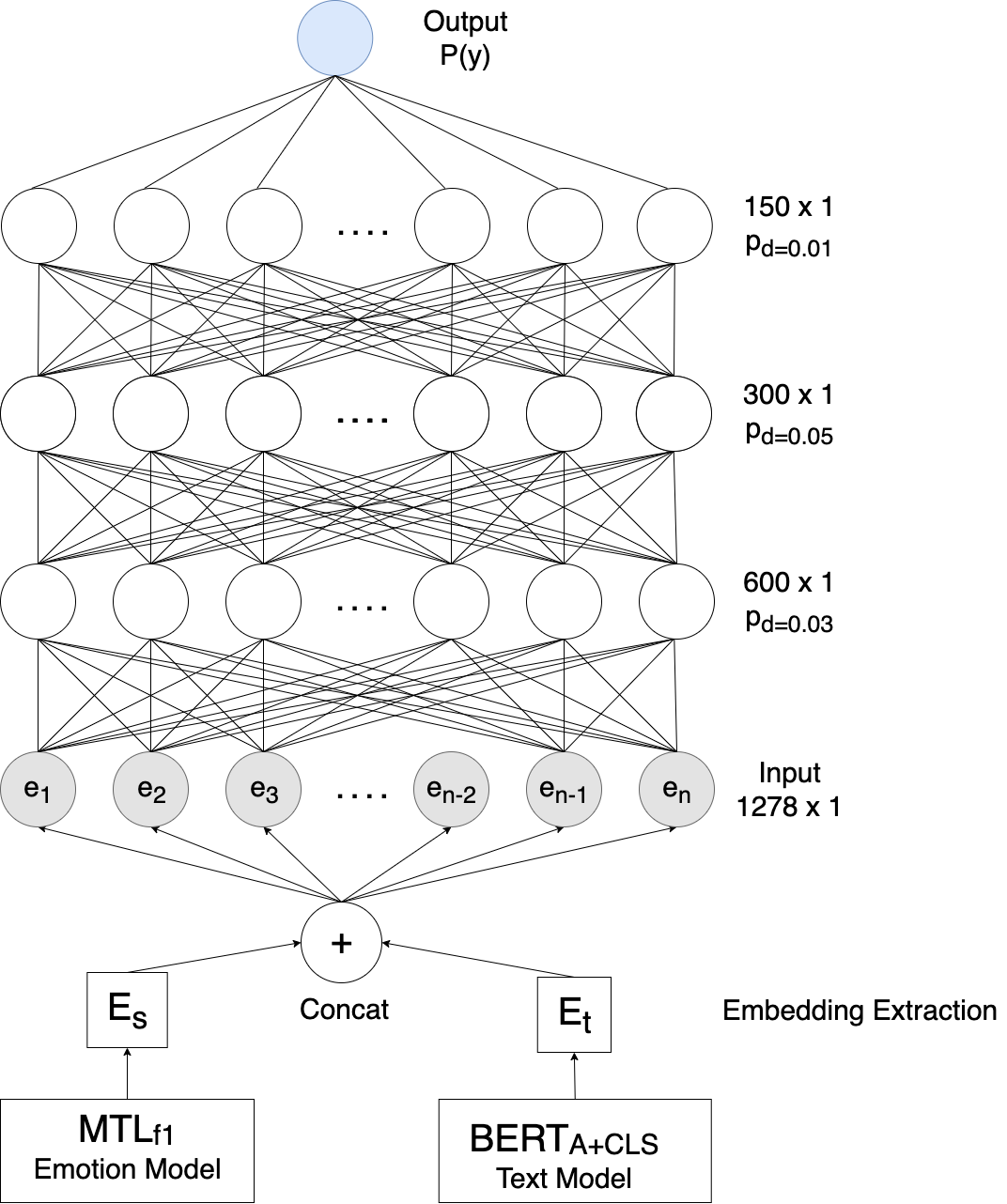}
\begin{figure}[!htbp]
\centering \makeatletter\IfFileExists{images/c1ee1a9e-507c-4c25-b71d-bc22fb87a69c-umml_model-udrawio.png}{\includegraphics[width=.82\linewidth]{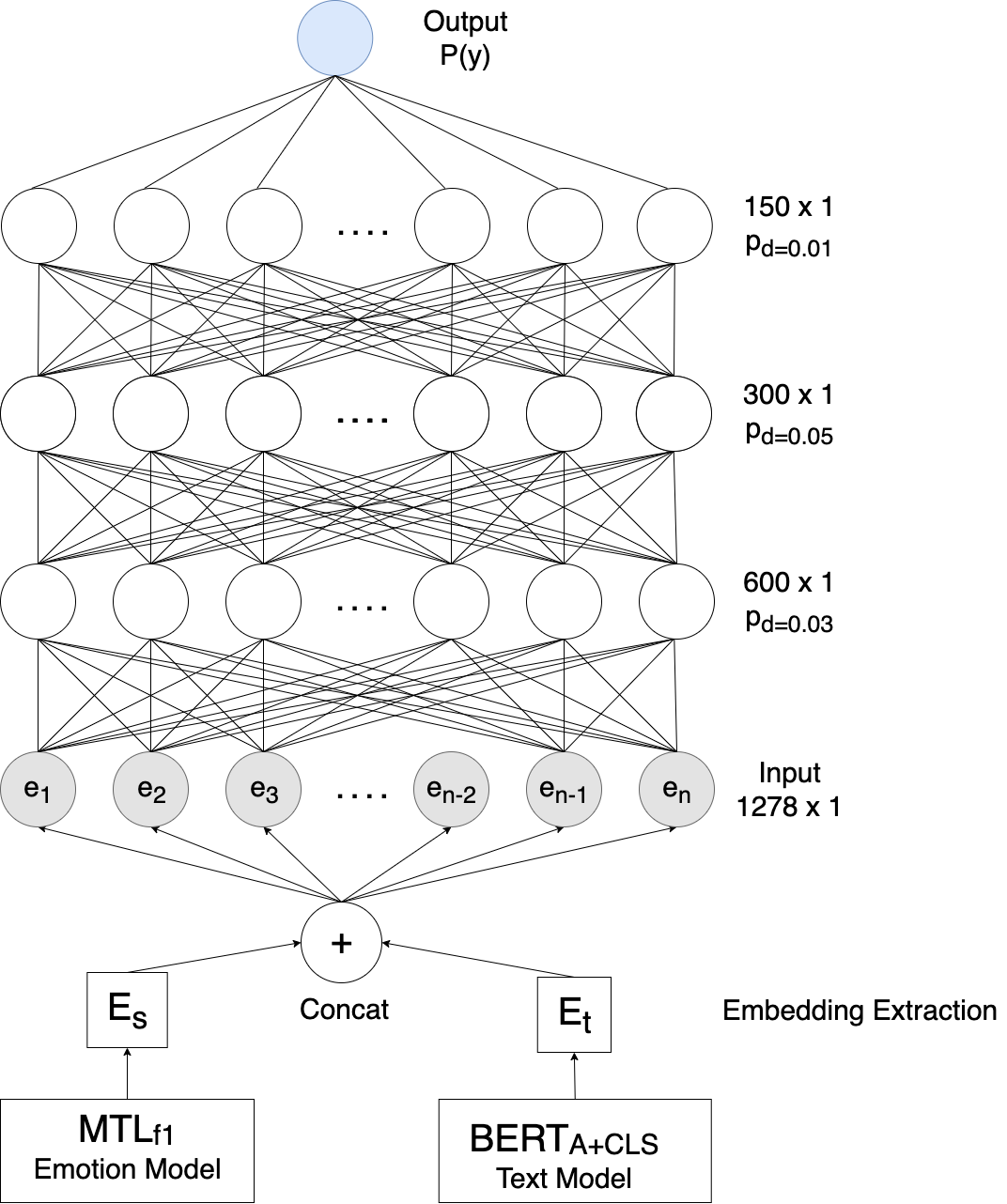}}{}
\makeatother 
\caption{{Multimodal learning model to predict the probability of hate speech for the input. MTL\ensuremath{_{f1}} and BERT\ensuremath{_{A+CLS}} are the pre-trained models after removing their respective final layers.}}
\label{f-00ad823cc8aa}
\end{figure}
\egroup
To optimize the model parameters a binary cross entropy loss is minimized. Additionally, for all the layers a \textit{L2 weight constraint }$\vert\vert w\vert\vert $ was applied to keep the learned weights small, as suggested by Hinton et al. \unskip~\cite{1298898:23922876}. Thus, the final loss function can be defined as,
\let\saveeqnno\theequation
\let\savefrac\frac
\def\dispfrac{\displaystyle\savefrac}
\begin{eqnarray}
\let\frac\dispfrac
\gdef\theequation{6}
\let\theHequation\theequation
\label{dfg-9752208f2ac3}
\begin{array}{@{}l}\begin{array}{l}L_{BCE}\;=\;-\frac1N\sum_{i=1}^{N}y_i\;.\;\log(p(y_i))\;\\\;\;\;\;\;\;\;\;\;\;\;\;\;\;\;+\;(1-y_i)\;.\;\log(1-p(y_i))\;+\;\alpha_w\vert\vert w\vert\vert_2\end{array}\end{array}
\end{eqnarray}
\global\let\theequation\saveeqnno
\addtocounter{equation}{-1}\ignorespaces 
where $N $ is the number of training examples, $\alpha_w $ is the hyperparameter that decides the amount of penalty on the weights, $y $ denotes the true labels and $p(y) $ denotes the predicted labels. The model was optimized during training using the Adam optimizer, which has been proven to outperform others especially towards the end of optimization as gradients become sparser\unskip~\cite{1298898:23923224}.
    
\section{\textbf{Experiments and Evaluation}}

\subsection{\textbf{Text Model Experiments}}Multiple combinations of datasets and models are experimented with to generate embeddings. Both BERT and ALBERT were pre-trained on all three A, B, and C, as shown in Table~\ref{tw-512fe3e1f581}. To select the best model for benchmark, they are fine-tuned on the HSDVD training set and evaluated on a holdout test set. The baseline models can be used to test the relative performance of multimodal learning when audio features are combined with the text features. The same train and test sets are used to evaluate the MML model so that the text model can be used as a true benchmark and ensure a fair comparison.

The input to both the models is a token sequence generated by breaking down sentences after pre-processing as specified in section 4(A). The maximum length of the sequence is limited to 128, and sequences smaller than this are padded with zeros as suggested by Devlin et al.\unskip~\cite{1298898:23922403}. Each model was trained for 6 epochs with batch size 8, learning rate 5e-6, and Adam optimizer with a weight decay rate of 0.01. After each epoch, the model was evaluated on a validation set, and the best performing epoch was saved.

Macro average of precision, recall, and f1-score of both the classes are used for an unbiased comparison. As seen from Table~\ref{tw-512fe3e1f581} both BERT\ensuremath{_{A}} and ALBERT\ensuremath{_{B}} models show competing results. Therefore, pre-trained versions of both models with the specified parameters will be used to generate text embeddings in the MML experiments.
\begin{table}[!htbp]
\caption{{Baseline Text Model Performance} }
\label{tw-512fe3e1f581}
\def\arraystretch{1}
\ignorespaces 
\centering 
\begin{tabulary}{\linewidth}{p{\dimexpr.25\linewidth-2\tabcolsep}p{\dimexpr.25\linewidth-2\tabcolsep}p{\dimexpr.25\linewidth-2\tabcolsep}p{\dimexpr.25\linewidth-2\tabcolsep}}
\hline Model &  P & R & F1\\
\hline 
BERT\ensuremath{_{A}} &
  \textbf{90.50} &
  91.73 &
  \textbf{91.11}\\
BERT\ensuremath{_{B}} &
  88.97 &
  91.12 &
  90.03\\
BERT\ensuremath{_{C}} &
  90.30 &
  89.00 &
  89.64\\
ALBERT\ensuremath{_{A}} &
  89.02 &
  91.25 &
  90.12\\
ALBERT\ensuremath{_{B}} &
  90.20 &
  \textbf{91.99} &
  91.08\\
ALBERTc &
  90.1 &
  88.5 &
  89.29\\
\hline 
\end{tabulary}\par 
\end{table}

\subsection{\textbf{Emotion Model Experiments}}The interactive emotional dyadic motion capture (IEMOCAP) dataset is used for training and evaluation of the emotion model. It consists of 10039 audio-visual recordings of actors carrying out scripted dialogue scenarios to express various emotional states. Since the models required fixed-size representation of input and the sample lengths vary from 1 to 34 seconds, we experimented with two different feature representations, f1 and f2. Both representations have similar features like energy, entropy, chroma, spectral and mel frequency cepstral coefficient (MFCC). However, the first set contains long term features calculated over the entire sample, and the second set consists of mid term features calculated for 10 seconds of the sample recording. As described in section 4(B), 68 short term features are calculated over a window size of 50 ms and step size of 50 ms at sampling frequency of 44kHz. Then 136 mid term features are calculated over the short term sequence, with window size and step size of 1000 ms. For f1, the long term features are further calculated by averaging over the mid term sequence. On the other hand, for f2, the mid term feature vectors are concatenated, padded with zeros, and truncated accordingly to obtain a fixed-size representation of 1360 features spanning 10 seconds of each input signal. The feature values for both the experiments are scaled down between -1 and 1 before being fed to the neural network.

\begin{table}[!htbp]
\caption{{Fine-tuned hyperparamters for the emotion model} }
\label{tw-1252c47e3e56}
\centering 
\begin{threeparttable}

\def\arraystretch{1}
\ignorespaces 
\centering 
\begin{tabulary}{\linewidth}{LLLLL}
\hline Model & Parameter & Value & Parameter & Value\\
\hline 
\multicolumn{1}{p{\dimexpr(\mcWidth{1})}}{\multirow{4}{\linewidth}{MTL\ensuremath{_{f1}}}} &
  epoch &
  30 &
  alpha &
  0.2\\
 &
  learning rate &
  1e-4 &
  beta &
  0.1\\
 &
  learning decay &
  0.99 &
  gamma &
  0.2\\
 &
  L2 regularization &
  1e-7 &
  batch size &
  32\\
\multicolumn{1}{p{\dimexpr(\mcWidth{1})}}{\multirow{4}{\linewidth}{MTL\ensuremath{_{f2}}}} &
  epoch &
  18 &
  alpha &
  0.1\\
 &
  learning rate &
  1e-3 &
  beta &
  0.1\\
 &
  learning decay &
  0.96 &
  gamma &
  0.2\\
 &
  L2 regularization &
  1e-9 &
  batch size &
  128\\
\hline 
\end{tabulary}\par 
\begin{tablenotes}\footnotesize 
    
\item{}
\end{tablenotes}
\end{threeparttable}

\end{table}
As seen in Figure~\ref{f-87efa4b75f31}, the MTL model shares two hidden layers. A dropout of 0.2 is applied to the output of these layers before passing to the attribute specific dense layer. All the hidden units are activated with ReLU function, and weights are initialized with He normalized distribution\unskip~\cite{1298898:23922897}. To further increase generalization, an L2 kernel regularization was also applied\unskip~\cite{1298898:23922876}. The hyperparameter for regularization is tuned using a validation dataset. Table~\ref{tw-1252c47e3e56} lists the model parameters that were tuned over a hyperband during training by using validation loss as the measure of performance. As specified in section 4(B), this also includes $\alpha,\;\beta,\;\gamma $ loss weights for valence, arousal, and dominance, respectively.

\begin{table}[!htbp]
\caption{{Emotion model performance} }
\label{tw-f51b68f568f8}
\def\arraystretch{1}
\ignorespaces 
\centering 
\begin{tabulary}{\linewidth}{LLLL}
\hline Model & Val RMSE & Aro RMSE & Dom RMSE\\
\hline 
MTL\ensuremath{_{f1}} &
  0.2594 &
  0.4232 &
  0.4418\\
MTL\ensuremath{_{f2}} &
  \textbf{0.1846} &
  \textbf{0.1124} &
  \textbf{0.1431}\\
\hline 
\end{tabulary}\par 
\end{table}
We evaluate the two regression models using \textit{Root Mean Squared Error (RMSE). }It is a measure of the standard deviation between the residuals i.e., the prediction errors. As seen in Table~\ref{tw-f51b68f568f8}, MTL\ensuremath{_{f2}} outperforms MTL\ensuremath{_{f1}} in prediction of all three emotion attributes and is finally used to generate speech embeddings for the MML model.

\begin{table*}[t!]
\caption{{Samples from HSDVD test set where superior performance of MML is observed.} }
\label{tw-d77d9ff08f7e}
\def\arraystretch{1}
\ignorespaces 
\centering 
\begin{tabulary}{\linewidth}{p{\dimexpr.7334\linewidth-2\tabcolsep}p{\dimexpr.13410000000000002\linewidth-2\tabcolsep}p{\dimexpr.13250000000000002\linewidth-2\tabcolsep}}
\hline Sample & BERT\ensuremath{_{Baseline}} & BERT\ensuremath{_{A+CLS}}\\
\hline 
Yea, if you are a black man, you better stay out of trouble and keep your goddamn hands outside of your pocket. You are basically begging to get shot in public. But if you are indiana nick bear drapped in a snake flag and some ninji star from the mall. Yeah, welcome to congress sir. \mbox{}\protect\newline  &
  HateSpeech &
  Not HateSpeech\\
The white british population has decreased by six hundered thousand. While the minority population has increased by one twenty million. Yes lads, we are winning.  \mbox{}\protect\newline  &
  HateSpeech &
  Not HateSpeech\\
Anita Hill testifies that supreme court nominie Clarence Thomas sexually harrased her. Hill was called a scorned women and a litle bit nutty and a litle bit sl*tty. \mbox{}\protect\newline  &
  HateSpeech &
  Not HateSpeech\\
Maternity flight suits, pregnant women are gonna fight our war. Its a mockery of the US military. China's military becomes more masculine, our military needs to become as joe biden says feminine, whatever feminine means anymore. &
  Not HateSpeech &
  HateSpeech\\
\hline 
\end{tabulary}\par 
\end{table*}

\subsection{\textbf{Multimodal Learning Experiments}}The MML model is trained on HSDVD consisting of 1k video recordings. These are first converted to wav audio formats using ffmpeg library\footnote{\protect\BreakURLText{https://ffmpeg.org/}} before speech processing for the emotion model. Since the data was collected from sources like Twitter and YouTube, it is more prone to noise; unlike the IEMOCAP dataset which was recorded in a controlled environment. We apply the spectral gating technique using Audacity\footnote{\protect\BreakURLText{https://www.audacityteam.org/}} for noise reduction. It is then fed to the MTL\ensuremath{_{f2}} model to generate speech embeddings E\ensuremath{_{s}} (Equation~(\ref{dfg-74fb9942c041})) of size 510, since each attribute specific layer consists of 170 hidden units.

For text processing, the speech samples are first converted to text using Houndify API\footnote{\protect\BreakURLText{https://www.houndify.com/}}. Further pre-processing steps are bypassed, as this data is not noisy like the Twitter data. It does contain some special characters like punctuation and contractions, which were preserved in the Twitter data during pre-training as well. The sentences are converted to token sequences of size 128 before processing by the selected text models i.e., BERT\ensuremath{_{A}} and ALBERT\ensuremath{_{B}}. A special token [CLS] is appended to the beginning of each sequence as given in section 4(A). The final hidden vector of this token can be used as a fixed size feature representation for the input sequence. It is the recommended method according to the original work\unskip~\cite{1298898:23922403}. Another well-known method used is averaging all the contextual word embeddings extracted from the last hidden layer (for example: \unskip~\cite{1298898:23924917,1298898:23924920,1298898:23924921}). These two options are also provided by the popular bert-as-a-service repository\footnote{\protect\BreakURLText{https://github.com/hanxiao/bert-as-service/}}, hence we used both methods in our experiments. The final feature vector generated using either method i.e., E\ensuremath{_{t}} has a size of 768.

\begin{table}[!htbp]
\caption{{MML Model Performance} }
\label{tw-6383d207e0fb}
\def\arraystretch{1}
\ignorespaces 
\centering 
\begin{tabulary}{\linewidth}{LLLL}
\hline Model & P & R & F1\\
\hline 
BERT\ensuremath{_{A+CLS}} &
  \textbf{93.00} &
  \textbf{92.89} &
  \textbf{92.94}\\
BERT\ensuremath{_{A+avg}} &
  90.60 &
  91.70 &
  91.14\\
BERT\ensuremath{_{Baseline}} &
  90.50 &
  91.73 &
  91.11\\
ALBERT\ensuremath{_{B+CLS}} &
  \textbf{92.36} &
  \textbf{92.90} &
  \textbf{92.62}\\
ALBERT\ensuremath{_{B+avg}} &
  90.34 &
  91.76 &
  91.04\\
ALBERT\ensuremath{_{Baseline}} &
  90.20 &
  91.99 &
  91.08\\
\hline 
\end{tabulary}\par 
\end{table}
The input embedding E obtained by concatenating E\ensuremath{_{s}} and E\ensuremath{_{t}} (Equation~(\ref{dfg-6299b0b1dd01})) consists of 1278 features. It is fed to the MML neural network depicted in Figure~\ref{f-00ad823cc8aa}. The dataset is split into training (80\%), validation (10\%), and testing (10\%). The validation set is used for tuning the hyperparameters such as learning rate of 1e-4 and decay rate of 0.99 for the Adam optimizer. Furthermore, early stopping is applied to stop model training if validation loss does not decrease for more than 10 epochs. As given in section 4(C) the binary cross entropy loss is calculated using Equation~(\ref{dfg-9752208f2ac3}).

For evaluation, similar metrics to baseline model are used i.e. macro average of precision, recall and f1-score of both the classes. Table~\ref{tw-6383d207e0fb}, lists the evaluation scores for both BERT\ensuremath{_{A}} and ALBERT\ensuremath{_{B}} with different text embedding extraction techniques. Baseline results have also been included for evaluation of the MML model performance. The BERT\ensuremath{_{A+CLS}} outperforms all the models in every metric, comparable to state-of-the art text models. Additionally, both BERT\ensuremath{_{A+CLS}} and ALBERT\ensuremath{_{B+CLS}} perform significantly better than their respective baselines, thereby indicating the significance of speech features when processing multimedia input. It should also be noted that embeddings extracted by averaging all token representations did not result in much increase in performance for BERT\ensuremath{_{A+avg}}. On the other hand the performance of ALBERT\ensuremath{_{B+avg}} slightly decreased as compared to its baseline. Therefore, the mean pooling technique might not produce a good representation of its input during such transfer learning tasks.

To further evaluate the functionality of MML we looked at some examples listed in Table~\ref{tw-d77d9ff08f7e} that were miss-classified by the BERT\ensuremath{_{Baseline}} model. Notably, most miss classifications were false positives, i.e., the content was classified as hate speech when it was not. This is in line with the majority increase in the precision of BERT\ensuremath{_{A+CLS}} compared to the increase in recall. For better intuition, consider samples 1 and 2, which were spoken sarcastically. At a glance, the keywords and wordings might indicate hate speech; however, the mocking tone of the speaker will clearly distinguish the utterance as sarcasm. Consider another sample 3, it was part of an informational segment on Twitter. It also consists of several sexist words such as \textit{scorned women, nutty and sl*tty, }but was correctly classified as not hate speech by the MML model, which can also be intrinsically attributed to the tone of the speaker. Rectifying false positives for such content which aims at spreading awareness about racism and sexism, becomes increasingly important to prevent stifling the freedom of speech in the process.
    
\section{\textbf{Conclusion}}
In this paper, we proposed a multimodal learning framework that considers both the tone and the words of the speaker in a video or audio to determine hate speech. The model benefits from the mutually beneficial relationship that exists between these modalities in the real world. Our experiments demonstrated that adding the speech features is more beneficial than just relying on the text features when detecting hate speech in the multimedia domain. This initial evidence suggests that including multiple other modalities and adding knowledge from different perspectives can further enhance the model's performance. For instance, we expect hate speech detection to also benefit from visual features such as detecting religious attires/objects, violence, or facial expressions. Finally, this also calls attention to the need for a different kind of system for hate speech detection in multimedia data which accounts for a large portion of the internet today and encourages new research avenues for this task.



%

\bibliographystyle{IEEEtran}

\bibliography{\jobname}

\end{document}